\documentclass[10pt,twocolumn,letterpaper]{article}

\usepackage[pagenumbers]{cvpr}      
\usepackage{multirow}
\usepackage{enumitem}
\usepackage{tikz}
\usetikzlibrary{arrows.meta,positioning,fit,backgrounds,calc}
\usepackage{microtype}
\usepackage[hidelinks,breaklinks=true]{hyperref}

\def\paperID{}
\def\confName{CVPR}
\def\confYear{2026}

\definecolor{qwenc}{RGB}{37,99,235}
\definecolor{clipc}{RGB}{22,163,74}
\definecolor{opc}{RGB}{217,119,6}
\definecolor{datac}{RGB}{100,116,139}

\newcommand{\Vr}{\ensuremath{V_r}}
\newcommand{\Ed}{\ensuremath{E}}
\newcommand{\Rk}[1]{\textnormal{R@}#1}

\title{Reason, Retrieve, Re-rank: A Zero-Shot Reasoning-Aware
Framework for Composed Video Retrieval}

\author{Ali Alavi\\
The Ohio State University\\
{\tt\small alavibajestan.1@osu.edu}}

\begin{document}
\maketitle

\begin{abstract}
\noindent
Composed Video Retrieval (CoVR) seeks the target video that results from
applying a free-form textual modification to a reference video. We address the
\emph{Reason-Aware} CoVR (CoVR-R) challenge at the CVPR~2026 VidLLMs workshop,
where retrieval is strictly zero-shot. We present \textbf{R3-CoVR}
(\emph{Reason, Retrieve, Re-rank}), a training-free pipeline built entirely
from frozen foundation models. A multimodal large language model
(Qwen3-VL-8B) reasons about the \emph{after-effects} an edit implies---state
transitions, action phases, scene, camera and tempo---and verbalises a concise
post-edit description; a contrastive video--text encoder (SigLIP-2) embeds this
description and the gallery for first-stage retrieval; finally a
constraint-aware re-ranking stage uses the same multimodal model as a judge
that scores each shortlisted candidate against the intended edited result.
On the challenge test set, R3-CoVR attains \textbf{91.9\% R@1} and
\textbf{98.2\% R@10}. Two findings drive these results: (i)~matching the
description length to the contrastive encoder's text window lifts \Rk{1} from
$67.5$ to $72.7$; and (ii)~the constraint-aware re-ranker, which reorders only
the shortlist, lifts \Rk{1} from $72.7$ to $91.9$---the single largest gain.
We analyse the re-ranker's behaviour, the retrieve/re-rank blend, and the
shortlist depth, and we release a clean three-layer implementation.
\end{abstract}

\section{Introduction}
Composed Video Retrieval (CoVR) generalises text-to-video retrieval with a
visual anchor: given a reference video \Vr{} and a modification instruction
\Ed, a system must retrieve, from a large gallery, the target video that
reflects the requested compositional change while preserving the relevant
context of \Vr~\cite{covr2024}. The task underpins content-based video
recommendation, intelligent editing, and creative search.

The CoVR-R challenge at the CVPR~2026 VidLLMs workshop sharpens the problem
along two axes. First, retrieval is \emph{zero-shot}: models may not be trained
or tuned on the challenge data, so the gallery must be matched by
generalisation rather than memorisation. Second, the benchmark is
\emph{reasoning-aware}: edits are written so that their visual consequences are
\emph{implied} rather than stated. ``Change typing to frustration'' implies
clenched fists and a closing laptop; ``make it a close-up'' implies tighter
framing and shorter duration. Hard distractors that share surface vocabulary
with the edit but violate its implied consequences defeat na\"ive keyword
matching~\cite{covrr2026}.

\begin{figure*}[t]
\centering
\resizebox{\textwidth}{!}{%
\begin{tikzpicture}[
  font=\footnotesize,
  box/.style={rounded corners=2.5pt, draw, align=center, inner sep=3.5pt, minimum height=8.5mm},
  qwen/.style={box, fill=qwenc!8, draw=qwenc!55, text=black},
  clip/.style={box, fill=clipc!10, draw=clipc!55},
  data/.style={box, fill=datac!12, draw=datac!60},
  op/.style={box, fill=opc!12, draw=opc!70},
  arr/.style={-{Latex[length=2mm]}, semithick, draw=black!70},
  lbl/.style={font=\scriptsize\itshape}
]
\node[data,text width=15mm] (vr)   at (0,1.65) {Reference\\ video $V_r$};
\node[data,text width=15mm] (ed)   at (0,0.55) {Edit text $E$};
\node[qwen,text width=30mm] (reason) at (3.55,1.1)
  {\textbf{Qwen3-VL} \textit{(frozen)}\\[1pt]\textit{after-effect reasoning}\\
   $R{=}\{$states, actions,\\ scene, camera, tempo$\}$};
\node[qwen,text width=21mm] (desc) at (7.05,1.1) {concise post-edit\\ description $D$};
\node[clip,text width=18mm] (qtext) at (9.75,1.1) {SigLIP-2\\ text enc.\ $\to q$};
\node[data,text width=16mm] (gal)  at (4.3,-1.35) {Gallery\\ videos $\{V_i\}$};
\node[clip,text width=22mm] (gvid) at (7.5,-1.35) {SigLIP-2 frames\\ mean-pool $\to v_i$};
\node[op,text width=16mm] (cos) at (12.4,-0.1) {cosine sim.\\ top-$K$ shortlist};
\node[qwen,text width=27mm] (judge) at (12.4,-2.6)
  {\textbf{Qwen3-VL} \textit{judge}\\[1pt]\textit{constraint score} $s_j$\\ (5-sample avg.)};
\node[op,text width=22mm] (fuse) at (15.7,-1.35)
  {blend\\ $\lambda s_j{+}(1{-}\lambda)\mathrm{rank}$};
\node[data,text width=14mm] (out) at (18.4,-1.35) {\textbf{final}\\ ranking};
\draw[arr] (vr.east)  -- (reason.west|-vr.east);
\draw[arr] (ed.east)  -- (reason.west|-ed.east);
\draw[arr] (reason) -- (desc);
\draw[arr] (desc) -- (qtext);
\draw[arr] (gal) -- (gvid);
\draw[arr] (qtext.east) -| (cos.north);          
\draw[arr] (gvid.east)  |- (cos.west);           
\draw[arr] (cos.south) -- node[lbl,right=1pt]{shortlist} (judge.north); 
\draw[arr] (judge.east) -| (fuse.south);         
\draw[arr] (cos.east) -| (fuse.north);           
\draw[arr] (fuse) -- (out);
\begin{scope}[on background layer]
\node[draw=qwenc!40,dashed,rounded corners,fit=(reason)(desc),inner sep=5pt,
      label={[lbl,text=qwenc!80]above:Stage 1: Reason}] {};
\node[draw=clipc!45,dashed,rounded corners,fit=(qtext)(gvid)(cos),inner sep=5pt,
      label={[lbl,text=clipc!80]above:Stage 2: Retrieve}] {};
\node[draw=opc!55,dashed,rounded corners,fit=(judge)(fuse)(out),inner sep=5pt,
      label={[lbl,text=opc!90]below:Stage 3: Re-rank}] {};
\end{scope}
\end{tikzpicture}}
\caption{\textbf{R3-CoVR.} A frozen multimodal LLM reasons about the
after-effects of the edit and writes a concise target description (Stage~1);
a frozen contrastive encoder embeds the description and the gallery and returns
a top-$K$ shortlist (Stage~2); the multimodal LLM then judges each shortlisted
candidate against the intended result, and the judge score is blended with the
retrieval rank to reorder the head of the list (Stage~3). No component is
trained on challenge data.}
\label{fig:pipeline}
\end{figure*}
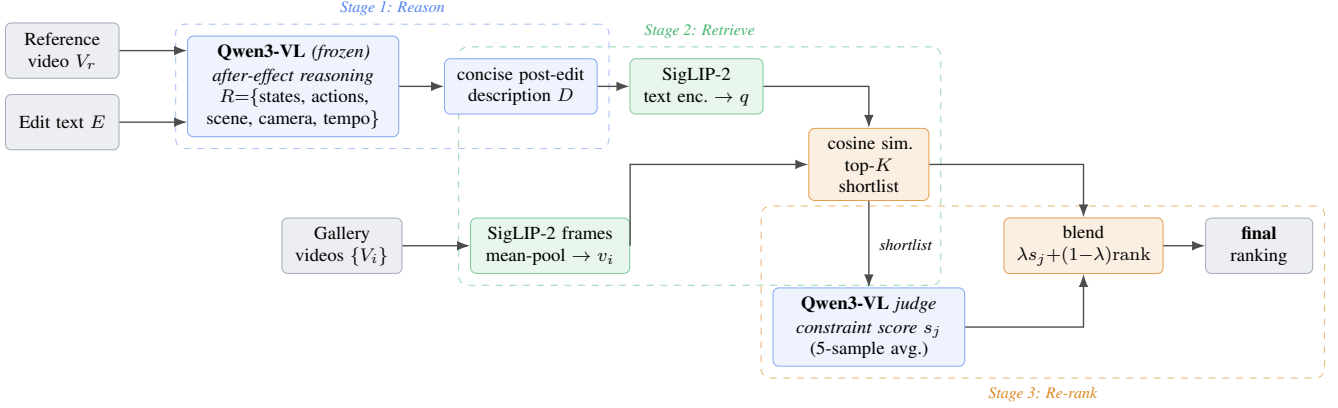

These constraints point away from trained fusion networks and toward a
\emph{verbalize-then-match} strategy, as established for composed \emph{image}
retrieval by training-free, language-driven methods such as
CIReVL~\cite{cirevl2024} and CoTMR~\cite{cotmr2025}. We adopt this view for
video and push it further with an explicit re-ranking stage. Our system,
\textbf{R3-CoVR}, has three stages (Fig.~\ref{fig:pipeline}):
\begin{enumerate}[leftmargin=1.3em,itemsep=1pt,topsep=2pt]
\item \textbf{Reason.} A frozen multimodal LLM watches \Vr{} and reasons about
the after-effects of \Ed{} as a structured trace, then writes a concise
post-edit target description.
\item \textbf{Retrieve.} A frozen contrastive video--text encoder embeds the
description and every gallery video; cosine similarity yields a shortlist.
\item \textbf{Re-rank.} The multimodal LLM acts as a constraint judge, scoring
each shortlisted candidate on how well it realises the intended result; scores
are fused with the retrieval rank to reorder the head of the list.
\end{enumerate}

Our contributions are: (1) a fully zero-shot, training-free CoVR pipeline that
attains $91.9\%$ \Rk{1} on the challenge test set; (2) the observation that
\emph{description length must match the contrastive text encoder's token
window}, worth $+5.2$ \Rk{1}; (3) a constraint-aware re-ranking stage that
contributes $+19.2$ \Rk{1}, with analysis of the retrieve/re-rank blend, the
shortlist depth, and the judge's score resolution; and (4) a reusable
three-layer (data / model / runner) implementation.

\section{Related Work}
\paragraph{Composed retrieval.}
Composed image retrieval (CIR) fuses a reference image with a modification
text and was popularised by CIRR and FashionIQ; CoVR extends the setting to
video, with WebVid-CoVR providing large-scale weakly-supervised
triplets~\cite{covr2024}. Trained approaches learn a cross-attention fusion
module, e.g.\ the enriched-context, discriminative-embedding line on
Dense-WebVid-CoVR~\cite{bsecovr2024}.

\paragraph{Training-free, language-driven retrieval.}
A complementary direction replaces the learned fusion module with language.
CIReVL~\cite{cirevl2024} captions the reference, lets an LLM rewrite the
caption under the modification, and retrieves with a frozen vision--language
encoder; CoTMR~\cite{cotmr2025} adds chain-of-thought, multi-scale reasoning.
These methods show that \emph{language is a strong compositional interface} and
that training-free pipelines are competitive. The CoVR-R
baseline~\cite{covrr2026} brings this to video with a frozen Qwen3-VL that
reasons about edit consequences before matching.

\paragraph{Re-ranking.}
Re-ranking a first-stage shortlist is standard in retrieval. In the
training-free composed setting, constraint-based re-rankers such as Soft
Filtering~\cite{soft2025} verify prescriptive and proscriptive textual
constraints to sharpen precision. R3-CoVR's Stage~3 is a multimodal
realisation of this idea: the judge inspects the candidate \emph{video} and
scores realisation of the edit, which is what most improves \Rk{1}.

\section{Method}
\subsection{Problem setup}
Given a reference video \Vr{} and a modification \Ed, rank a gallery
$\mathcal{G}=\{V_1,\dots,V_N\}$ so the ground-truth target $V^\star$ appears as
high as possible. We never train on challenge data; every model is frozen.

\subsection{Stage 1 -- Reason}
A frozen Qwen3-VL-8B observes \Vr{} (sampled at $1$\,fps) and is prompted to
emit a structured \emph{after-effect} trace
$R=\{\textsf{states},\textsf{actions},\textsf{scene},\textsf{camera},\textsf{tempo}\}$
followed by a target description $D$, in a single generation. Reasoning about
consequences---rather than copying edit keywords---is what separates the target
from hard distractors. To stabilise the description we draw $K{=}3$ samples
(self-consistency) and average their embeddings downstream. The trace also
serves as the human-legible \emph{reasoning trace} required by the challenge.

\subsection{Stage 2 -- Retrieve}
A frozen SigLIP-2 contrastive encoder maps text and video into a shared metric
space. Each gallery video is encoded from $8$ uniformly sampled frames,
mean-pooled and $L_2$-normalised; the query embedding $q$ is the normalised
mean of the description embeddings. Retrieval ranks $\mathcal{G}$ by cosine
similarity $s(V)=q^{\!\top} v(V)$, excluding the query's own reference, and
returns a top-$K$ shortlist.

\paragraph{Description length matters.}
The contrastive text encoder has a fixed token window (64 tokens for SigLIP-2).
A vivid, paragraph-length description is truncated, discarding salient
post-edit content. Constraining $D$ to a \emph{concise, front-loaded} caption
($\le 40$ words) that names the changed subjects, objects and setting fits the
window and is the single most effective Stage-2 choice
(\S\ref{sec:results}). The rich reasoning trace is retained separately for the
challenge's reasoning metric.

\subsection{Stage 3 -- Re-rank}
Stage~2 reliably places the target in the shortlist (\Rk{10}${\approx}97\%$)
but not always at rank~1. Stage~3 reorders the shortlist with the same frozen
Qwen3-VL acting as a \emph{constraint judge}: shown a candidate video, the edit
\Ed, and the intended description $D$, it returns an integer match score in
$[0,100]$, normalised to $s_j\in[0,1]$. The final score blends the judge with
the retrieval rank,
\begin{equation}
s_\text{final}(V) = \lambda\, s_j(V) + (1-\lambda)\,\mathrm{rank}_K(V),
\label{eq:blend}
\end{equation}
where $\mathrm{rank}_K(V)=(K-i)/K$ for shortlist position $i$. The judge is
discriminative---most distractors receive $0$ while true matches cluster near
$1$---so the blend with the retrieval rank breaks ties and guards against the
occasional false positive. Re-ranking touches only the $K$ shortlisted
candidates, bounding cost.

\paragraph{De-noising the judge.}
A greedy judge emits only a handful of distinct scores, so a target and a
distractor frequently tie at the maximum and the tiebreak decides rank~1. We
average the score over $5$ sampled generations, increasing the number of
distinct score values by an order of magnitude ($8\!\rightarrow\!86$ in our
test pool) and resolving these ties.

\section{Experiments}
\subsection{Dataset, metrics, implementation}
The CoVR-R challenge spans two galleries: WebVid (2{,}679 clips) and
Something-Something-v2 (1{,}654 clips). The test split has $301$ queries
($4$ WebVid, $297$ SS2); each query is ranked against all videos in its
gallery. We report Recall@$K$ for $K\in\{1,5,10,50\}$, the official metric;
the headline is \Rk{1}. Stage~1/3 use Qwen3-VL-8B-Instruct; Stage~2 uses
\texttt{siglip2-so400m-patch16-naflex}. All models are frozen and run on a
single A100~(40\,GB). Gallery embeddings are computed once and cached; the
shortlist depth is $K{=}10$ unless noted, and the blend is $\lambda{=}0.8$.

\subsection{Main results}
\label{sec:results}
Table~\ref{tab:main} traces the contribution of each design choice on the
challenge test set. First-stage retrieval with a rich description already
clears the published reference point, but truncation caps recall. Matching the
description to the encoder window (concise) lifts \Rk{1} by $+5.2$ and \Rk{5}
by a striking $+16.5$. The constraint-aware re-ranker then lifts \Rk{1} by a
further $+19.2$ to $\mathbf{91.9}$. Extending the shortlist to $K{=}20$ trades
a hair of \Rk{1} for the best broad recall ($\Rk{10}=98.2$), as the judge
promotes targets that first-stage retrieval buried below rank~10. For context,
the published CoVR-R reference reports $49.9$ \Rk{1} on its own benchmark
split~\cite{covrr2026}; our numbers are on the challenge test split and are not
a direct head-to-head, but the gap indicates the strength of the pipeline.

\begin{table*}[t]
\centering
\small
\caption{\textbf{Main results on the CoVR-R challenge test set} (301 queries).
Each row adds one design choice over the row above. Re-ranking uses the
Qwen3-VL constraint judge with blend $\lambda{=}0.8$.}
\label{tab:main}
\begin{tabular}{llcccc}
\toprule
& Configuration & \Rk{1} & \Rk{5} & \Rk{10} & \Rk{50}\\
\midrule
(a) & Reason\,+\,Retrieve, rich description            & 67.47 & 78.41 & 80.60 & 97.65\\
(b) & Reason\,+\,Retrieve, concise description          & 72.69 & 94.95 & 96.80 & 99.16\\
(c) & \;\;+\,constraint re-rank (top-10)                & \textbf{91.92} & 96.80 & 96.80 & 99.16\\
(d) & \;\;+\,constraint re-rank (top-20)                & 91.58 & \textbf{97.81} & \textbf{98.15} & \textbf{99.16}\\
\bottomrule
\end{tabular}
\end{table*}

\subsection{Ablations}
\paragraph{Retrieve/re-rank blend.}
Table~\ref{tab:blend} sweeps $\lambda$ in Eq.~\eqref{eq:blend} at $K{=}10$.
Pure retrieval ($\lambda{=}0$) is the Stage-2 result ($72.7$); pure judge
($\lambda{=}1$) reaches $88.2$ but discards the useful retrieval prior, while a
judge-dominant blend ($\lambda\!\in\![0.5,0.8]$) is best, peaking at $91.9$.
The judge supplies most of the signal, but retaining a fraction of the
retrieval rank guards against judge false positives.

\paragraph{Shortlist depth.}
Table~\ref{tab:depth} contrasts $K{=}10$ and $K{=}20$. A deeper shortlist gives
the judge access to targets ranked $11$--$20$ by Stage~2, improving \Rk{5} and
\Rk{10} ($96.8\!\to\!98.2$); \Rk{1} is essentially unchanged, as deeper lists
also expose more high-scoring distractors. Depth thus trades top-1 precision
for broad recall.

\paragraph{Judge resolution.}
Replacing the greedy judge with the $5$-sample averaged judge raises the number
of distinct candidate scores from $8$ to $86$ in the test pool, removing the
ties that otherwise force the retrieval-rank tiebreak to decide rank~1; this is
the mechanism the blend in Table~\ref{tab:blend} partially compensates for.

\begin{table}[ht]
\centering
\small
\caption{\textbf{Retrieve/re-rank blend} $\lambda$ (Eq.~\ref{eq:blend}),
top-10 shortlist, challenge test \Rk{1}. $\lambda{=}0$ is pure retrieval,
$\lambda{=}1$ is pure judge.}
\label{tab:blend}
\resizebox{\columnwidth}{!}{%
\begin{tabular}{lccccccc}
\toprule
$\lambda$ & 0.0 & 0.3 & 0.5 & 0.6 & 0.7 & \textbf{0.8} & 1.0\\
\midrule
\Rk{1} & 72.69 & 89.90 & 91.08 & 91.58 & 91.41 & \textbf{91.92} & 88.22\\
\bottomrule
\end{tabular}}
\end{table}

\begin{table}[ht]
\centering
\small
\caption{\textbf{Shortlist depth} for the constraint re-ranker
(blend $\lambda{=}0.7$). Deeper shortlists favour broad recall.}
\label{tab:depth}
\begin{tabular}{lcccc}
\toprule
Depth & \Rk{1} & \Rk{5} & \Rk{10} & \Rk{50}\\
\midrule
top-10 & \textbf{91.41} & 96.80 & 96.80 & 99.16\\
top-20 & 91.58 & \textbf{97.81} & \textbf{98.15} & 99.16\\
\bottomrule
\end{tabular}
\end{table}

\subsection{Implementation notes}
The system is organised in three layers---\emph{data}, \emph{model}, and
\emph{runner}---behind stable interfaces, with factory-instantiated encoders
and reasoners, so a backbone can be swapped without touching the orchestration.
Gallery embeddings are cached once per encoder; per-query reasoning and judge
scores are cached and resumable. First-stage retrieval over both galleries and
reasoning over all $301$ queries complete in well under an hour on one A100;
re-ranking the top-10 adds ${\sim}2$\,s per query.

\subsection{Discussion and limitations}
Three properties make R3-CoVR practical. It is \emph{training-free}: every
component is a frozen, publicly available model, so there is no risk of
over-fitting to the challenge data and the pipeline transfers directly to new
galleries. It is \emph{interpretable}: the Stage-1 trace and the Stage-3 judge
rationale expose \emph{why} a video is preferred, which the challenge also
rewards through its reasoning-trace metric. And it is \emph{modular}: stronger
encoders or judges drop in behind the factory interface.

Several limitations point to future work. (i)~The re-ranker's cost grows
linearly with the shortlist depth $K$ and with the judge's per-candidate
latency; very large galleries would benefit from a cheaper first pass or a
cascaded judge. (ii)~The judge inherits the backbone's perceptual biases; the
residual gap between \Rk{1} ($91.9$) and the \Rk{10} ceiling ($96.8$--$98.2$)
is dominated by a handful of fine-grained pairs that an $8$B judge scores
identically, and a larger judge (e.g.\ a $72$B variant) is the most direct way
to close it. (iii)~The contrastive encoder's $64$-token text window bounds how
much of the reasoned description can inform first-stage retrieval; a
long-context or video-native retriever would relax this and likely lift the
shortlist recall that upper-bounds the re-ranker. (iv)~Self-consistency and the
de-noised judge trade compute (3$\times$ generation, 5$\times$ scoring) for
robustness; the trade-off is favourable here but should be revisited at scale.

\section{Conclusion}
R3-CoVR shows that reason-aware composed video retrieval is well served by a
fully zero-shot, three-stage pipeline of frozen models: reason about the
edit's after-effects in language, retrieve with a contrastive encoder, and
re-rank the shortlist with the same multimodal model used as a constraint
judge. Two simple but consequential choices---fitting the description to the
encoder's token window, and re-ranking only the shortlist---take \Rk{1} from
$67.5$ to $91.9$ on the challenge test set, with the re-ranker contributing the
majority of the gain. The approach trains nothing, exposes interpretable
reasoning traces, and is straightforward to reproduce.



\begin{thebibliography}{9}\small
\bibitem{covrr2026}
O.~Thawakar, D.~Demidov, V.~Potlapalli, et al.
\newblock CoVR-R: Reason-Aware Composed Video Retrieval.
\newblock {\em CVPR (Findings)}, 2026. arXiv:2603.20190.

\bibitem{covr2024}
L.~Ventura, A.~Yang, C.~Schmid, G.~Varol.
\newblock CoVR: Learning Composed Video Retrieval from Web Video Captions.
\newblock {\em AAAI}, 2024.

\bibitem{bsecovr2024}
Composed Video Retrieval via Enriched Context and Discriminative Embeddings.
\newblock {\em CVPR}, 2024.

\bibitem{cirevl2024}
S.~Karthik, K.~Roth, M.~Mancini, Z.~Akata.
\newblock Vision-by-Language for Training-Free Compositional Image Retrieval.
\newblock {\em ICLR}, 2024.

\bibitem{cotmr2025}
CoTMR: Chain-of-Thought Multi-Scale Reasoning for Training-Free Zero-Shot
Composed Image Retrieval.
\newblock arXiv:2502.20826, 2025.

\bibitem{soft2025}
Soft Filtering: Guiding Zero-shot Composed Image Retrieval with Prescriptive
and Proscriptive Constraints.
\newblock arXiv:2512.20781, 2025.
\end{thebibliography}
\end{document}